\documentclass[11pt,a4paper]{article}
\usepackage[hyperref]{emnlp-ijcnlp-2019}
\usepackage{times}
\usepackage{latexsym}

\usepackage{url}

\usepackage{amssymb}
\usepackage{booktabs} 
\usepackage{enumitem}
\usepackage{multirow}
\usepackage{graphicx}
\usepackage{subfigure}

\aclfinalcopy %

\newif\ifdraft
\draftfalse

\usepackage{siunitx}

\newcommand*{\bleu}[1]{\num[round-mode=places,round-precision=1]{#1}}

\newcommand{\Ni}{({\em i})~}
\newcommand{\Nii}{({\em ii})~}
\newcommand{\Niii}{({\em iii})~}
\newcommand{\Niv}{({\em iv})~}
\newcommand{\Nv}{({\em v})~}

\newcommand{\SiEn}{Sinhala--English~}
\newcommand{\RSiEn}{English--Sinhala~}
\newcommand{\sien}{si--en~}

\newcommand{\NeEn}{Nepali--English~}
\newcommand{\RNeEn}{English--Nepali~}
\newcommand{\neen}{ne--en~}

\newcommand{\En}{English~}
\newcommand{\Si}{Sinhala~}
\newcommand{\Ne}{Nepali~}

\newcommand{\Sins}{Sinhala}
\newcommand{\Nens}{Nepali}
\newcommand{\HiNeEn}{\{Hindi, Nepali\}--English~}
\newcommand{\RHiNeEn}{English--\{Hindi, Nepali\}}
\newcommand{\gback}{\hspace{-0.5ex}}
\newcommand{\SiEnns}{Sinhala--English}
\newcommand{\NeEnns}{Nepali--English}
\newcommand{\RSiEnns}{English--Sinhala}
\newcommand{\RNeEnns}{English--Nepali}
\newcommand{\flores}{\textsc{Flores}}

\title{The \flores{} Evaluation Datasets for Low-Resource Machine Translation: \\
\NeEn and \SiEn}

\author{
Francisco Guzm\'{a}n$^{\heartsuit\blacklozenge}$ \enskip Peng-Jen Chen$^{\heartsuit\bigstar}$  \enskip Myle Ott$^{\bigstar}$ \enskip Juan Pino$^{\blacklozenge}$ \enskip Guillaume Lample$^{\bigstar\ddagger}$ \\
{\bf Philipp Koehn}$^{\blacksquare}$ \enskip {\bf Vishrav Chaudhary}$^{\blacklozenge}$ \enskip {\bf Marc'Aurelio Ranzato}$^{\bigstar}$ \\
$^{\blacklozenge}$Facebook Applied Machine Learning  \enskip $^{\bigstar}$Facebook AI Research  \\
\enskip $^{\ddagger}$Sorbonne Universit\'es \enskip $^{\blacksquare}$Johns Hopkins University  \\
\texttt{\{fguzman,pipibjc,myleott,juancarabina,guismay,vishrav,ranzato\}@fb.com} \enskip \texttt{phi@jhu.edu}\\
}

\date{}

\begin{document}
\maketitle

\begin{abstract}
For machine translation, %
a vast majority of language pairs in the world are considered low-resource because they have little %
parallel data available. %
Besides the technical challenges of learning with limited supervision, %
it is %
difficult to {\em evaluate} methods trained on low-resource language pairs because of the lack of %
freely and publicly available benchmarks. %
In this work, we introduce the \flores{} evaluation datasets for %
\NeEn and \SiEnns, based on sentences translated from Wikipedia. Compared to English, these are languages with very different morphology and syntax, for which little out-of-domain parallel data is available and for which relatively large amounts of monolingual data are freely available. We describe our process to collect and cross-check the quality of translations, and we report baseline performance using several learning settings: fully supervised, weakly supervised,
semi-supervised, and fully unsupervised. Our experiments demonstrate that current state-of-the-art methods perform rather poorly on this benchmark, posing a challenge to the research community working on low-resource MT. Data and code to reproduce our experiments are available at {\small \url{https://github.com/facebookresearch/flores}}.
\end{abstract}

\renewcommand{\thefootnote}{$^{\heartsuit}$}
\footnotetext[1]{Equal contribution.}
\renewcommand\thefootnote{\arabic{footnote}}

\section{Introduction} \label{sec:intro}
Research in Machine Translation (MT) has seen significant advances in recent years thanks to improvements in modeling,
and in particular neural 
models~\cite{sutskever2014sequence,attentionNMT,gehring2016convolutional,vaswani2017attention},
as well as the availability of large parallel corpora for training~\cite{TIEDEMANN12.463,smith2013dirt,bojar2017findings}.
Indeed, modern neural MT systems can achieve near human-level translation performance on language pairs
for which sufficient parallel training resources exist (e.g., Chinese--English translation~\cite{msrMT18} and English--French translation~\cite{gehring2016convolutional,ott2018analyzing}.

Unfortunately, MT systems, and in particular neural models, perform poorly on
\emph{low-resource} language pairs, for which parallel training data is scarce \cite{koehn17}. Improving translation performance on low-resource language pairs could be very impactful %
considering that these languages are spoken by a large fraction of the world population. 

Technically, there are several challenges to solve in order to improve translation for low-resource languages. First, in face of the scarcity of clean parallel data, MT systems should be able to use any source of data available, namely monolingual resources, noisy comparable data, as well as parallel data in related languages. Second, we need reliable public evaluation benchmarks to track progress in translation quality. 

Building evaluation sets on low-resource languages is both expensive and time-consuming because the pool of professional translators is limited, as there are few fluent bilingual speakers for these languages. Moreover, the quality of professional translations for low-resource languages is not on par with that of high-resource languages, given that the quality assurance processes for the low-resource languages are often lacking or under development. Also, it is difficult to verify the quality of the human translations as an non-native speaker, because %
the topics of the documents in these low-resource languages may require knowledge and context coming from the local culture.

\noindent

In this work, we introduce new evaluation benchmarks on two very low-resource language pairs: \NeEn and \SiEn\gback. Sentences were  extracted from Wikipedia articles in each language and translated by professional translators.
The datasets we release to the community are composed of a tune set of 2559 and 2898 sentences, a development set of 2835 and 2766 sentences, and a test set of 2924 and 2905 sentences for \NeEn and \SiEn respectively.

In \textsection\ref{sec:methodology}, we describe the methodology we used to collect the data as well as to check the quality of translations. 
The experiments reported in \textsection\ref{sec:experiments} demonstrate that these benchmarks are very challenging for current state-of-the-art methods, yielding very low BLEU scores~\cite{bleu} even using all available parallel data as well as monolingual data or Paracrawl\footnote{\url{https://paracrawl.eu/}} filtered data. This suggests that these languages and evaluation benchmarks can constitute a useful test-bed for developing and comparing MT systems for low-resource language pairs.

\section{Related Work}
There is ample literature on low-resource MT. From the modeling side, one possibility is to design methods that make more effective use of monolingual data.  This is a research avenue that has seen a recent surge of interest, starting with semi\-supervised methods relying on back-translation~\cite{sennrich2015improving}, integration of a language model into the decoder~\cite{gulcehre2017integrating, W18-6321} all the way to fully unsupervised approaches~\cite{lample_emnlp2018, artetxe_emnlp2018}, which use monolingual data both for learning good language models and for fantasizing parallel data. Another avenue of research has been to extend the traditional supervised learning setting to a {\em weakly supervised} one, whereby the original training set is augmented with parallel sentences mined from noisy comparable corpora like Paracrawl. In addition to the challenge of learning with limited supervision, low-resource language pairs often involve distant languages that do not share the same alphabet, or have very different morphology and syntax; accordingly, %
recent work has begun to explore language-independent lexical representations to improve transfer learning~\cite{gu2018meta}.
\noindent

In terms of low-resource datasets, DARPA programs like LORELEI~\cite{lorelei} have collected translations on several low-resource languages like English--Tagalog. Unfortunately, the data is only made available to the program's participants.
More recently, the Asian Language Treebank project~\cite{riza2016introduction} has introduced parallel datasets for several low-resource language pairs, but these are sampled from text originating in English and thus may not generalize to text sampled from low-resource languages.

In the past, there has been work on extracting high quality translations from crowd-sourced workers using automatic methods \cite{zaidan-callisonburch:2011:ACL-HLT2011, post-callisonburch-osborne:2012:WMT}. However, crowd-sourced translations have generally  lower quality than professional translations. In contrast, in this work we explore the quality checks that are required to filter \emph{professional} translations of low-resource languages in order to build a high quality benchmark set. %

In practice, there are very few publicly available datasets for low-resource language pairs, and often times, researchers {\em simulate} learning on low-resource languages by using a high-resource language pair like English--French, and merely limiting how much labeled data they use for training~\cite{gmt17,unsupNMTlample}. While this practice enables a framework for easy comparison of different approaches, the real practical implications deriving from these methods can be unclear. For instance, low-resource languages are often distant and often times corresponding corpora are not comparable, conditions which are far from the simulation with high-resource European languages, as has been recently pointed out by \citet{neubig18emnlp}.

\section{Methodology \& Resulting Datasets} \label{sec:methodology}
For the construction of our benchmark sets we chose to translate between Nepali and \Si into and out of English. Both Nepali and Sinhala are Indo-Aryan languages with a subject-object-verb (SOV) structure.  Nepali is similar to Hindi in its structure, while \Si is characterized by extensive omissions of arguments in a sentence.

Nepali is spoken by about 20 million people if we consider only Nepal, while Sinhala is spoken by about 17 million people just in Sri Lanka\footnote{See \url{https://www.ethnologue.com/language/npi} and \url{https://www.ethnologue.com/language/sin}.}. 
\noindent
\Si and Nepali have very little publicly available parallel data . For instance, most of the parallel corpora for \NeEn originate from GNOME and Ubuntu handbooks, and account for about $500$K sentence pairs.\footnote{Nepali has also $4$K sentences translated from English Penn Tree Bank at \url{http://www.cle.org.pk/software/ling_resources/UrduNepaliEnglishParallelCorpus.htm}, which is %
valuable parallel data.}
For \SiEnns, there are an additional $600$K sentence pairs automatically aligned from from OpenSubtitles~\cite{LisonTK18}.
Overall, the domains and quantity of the existing parallel data are very limited. However, both languages have a rather large amount of monolingual data publicly available \cite{CCNgram14}, making them perfect candidates to track performance on unsupervised and semi-supervised tasks for Machine Translation. %

\subsection{Document selection}
To build the evaluation sets, we selected and professionally translated sentences originating from Wikipedia articles in English, Nepali and \Si %
from a  %
Wikipedia snapshot of early May 2018. %
To select sentences for translation, we first selected the top $25$ documents that contain the largest number of \emph{candidate} sentences in each source language. %
To this end, we defined candidate sentences\footnote{We first used HTML markup to split document text into paragraphs. We then used regular expressions to split on punctuation, e.g. full-stop, poorna virama (\textbackslash u0964) and exclamation marks.} as: \Ni being in the intended source language according to a language-id classifier~\cite{bojanowski2016enriching}\footnote{This is a necessary step as many sentences in foreign language Wikipedias may be in English or other languages.}, and \Nii having sentences between $50$ and $150$ characters. %
Moreover, we considered sentences and documents
to be inadequate for translation when they contained large portions of untranslatable content such as lists of entities\footnote{For example, the Academy Awards page: \url{https://en.wikipedia.org/wiki/Academy_Award_for_Best_Supporting_Actor}.}. 
To avoid such lists we used the following rules: \Ni for English, sentences have to start with an uppercase letter and end with a period; \Nii for Nepali and Sinhala, sentences should not contain symbols such as bullet points, repeated dashes, repeated periods or ASCII characters.
The document set, along with the categories of documents is presented in the Appendix, Table~\ref{tab:wiki_docs}.\vspace{15pt}

\noindent
After the document selection process, we randomly sampled 2,500 sentences for each language. From English, we translated into \Ne and \Sins, while from \Si and \Nens, we only translated into English. We requested each string to be translated twice by different translators.

\subsection{Quality checks} \label{sec:checks}
Translating domain-specialized content such as Wikipedia articles from and to low-resource languages is challenging: the pool of available translators is limited, there is limited context available to each translator when translating one string at a time, and some of the sentences can contain code-switching (e.g. text about Buddhism in \Ne or \Si can contain Sanskrit or Pali words). As a result, we observed large variations in the level of translation quality, which motivated us to enact a series of automatic and manual checks to filter out poor translations.

We first used automatic methods to filter out poor translations and sent them for rework. Once the reworked translations were received, we sent all translations (original or reworked) that passed the automatic checks to human quality checks. Translations which failed human checks, were disregarded. Only the translations that passed all checks were added to the evaluation benchmark, although some source sentences may have less than two translations.
Below, we describe the automatic and manual quality checks that we applied to the datasets.

\paragraph{Automatic Filtering.} The guiding principles underlying our choice of automatic filters are: \Ni translations should be fluent \cite{zaidan-callisonburch:2011:ACL-HLT2011}, \Nii they should be sufficiently different from the source text, \Niii translations should be similar to each other, yet not equal; and \Niv translations should not be transliterations.
In order to identify the vast majority of translation issues we filtered by: \Ni applying a count-based n-gram language model trained on Wikipedia monolingual data and removing translations that have perplexity above 3000.0 (English translations only), \Nii  removing translations that have sentence-level char-BLEU score between the two generated translations below 15 (indicating disparate translations) or above 90 (indicating suspiciously similar translations),
\Niii removing sentences that contain at least 33\% transliterated words, \Niv removing translations where at least 50\% of words are copied from the source sentence, %
 and \Nv removing translations that contain more than 50\%  out-of-vocabulary words or more than 5 total out-of-vocabulary words in the sentences (English translations only). For this, the vocabulary was calculated on the monolingual English Wikipedia described in Table~\ref{tab:parallel-data}.
 \ifdraft{We also removed translations that have sentence-level word-BLEU score above 95 to the output of publicly available commercial translation services.}
 \fi

\paragraph{Manual Filtering.} 
We followed a setup similar to \emph{direct assessment} \cite{graham-EtAl:2013:LAW7-ID}. We asked three different raters to rate sentences from 0--100 according to the perceived translation quality. In our guidelines, the 0--10 range represents a translation that is completely incorrect and inaccurate, the 70--90 range represents a translation 
that closely preserves the semantics of the source sentence,  while the 90--100 range represents a \emph{perfect} translation. To ensure rating consistency, we rejected any evaluation set in which the range of scores among the three reviewers was above 30 points, and requested a fourth rater to break ties, by replacing the most diverging translation rating with the new one. 
For each translation, we took the average score over all raters and rejected translations whose scores were below 70. 

To ensure that the translations were as fluent as possible, we also designed an Amazon Mechanical Turk (AMT) monolingual task to judge the {\em fluency} of English translations. %
Regardless of content preservation, translations that are not fluent in the target language should be disregarded.
For this task, we then asked five independent human annotators to rate the fluency of each English translation from 1 (bad) to 5 (excellent), and retained only those above 3.%
 Additional statistics of automatic and manual filtering stages can be found in Appendix.

\subsection{Resulting Datasets}

 We built three evaluation sets for each language pair using the data that passed our automatic and manual quality checks: \emph{dev}~(tune), \emph{devtest}~(validation) and \emph{test}~(test).  The tune set is used for hyper-parameter tuning and model selection, the validation set is used to measure generalization during development, while the test set is used for the final blind evaluation.

To measure performance in both directions (e.g. \SiEn and \RSiEnns), we built test sets with mixed original-\emph{translationese} \cite{baroni2005new} on the source side. To reduce the effect of the source language on the quality of the resulting evaluation benchmark, direct %
and reverse translations %
were mixed at an approximate 50-50 ratio for the \emph{devtest} and \emph{test} sets. On the other hand, the \emph{dev} set was composed of the remainder of the available translations, which were not guaranteed to be balanced. Before selection, the sentences were grouped by document, to minimize the number of documents per evaluation set. 

In Table~\ref{tab:set_stats} we present the statistics of the resulting sets. For Sinhala--English, the \emph{test} set is composed of 850 sentences originally in English, and 850 originally in Sinhala. We have approximately 1.7 translations per sentence. This yielded 1,465 sentence pairs originally in English, and  1,440 originally in Sinhalese, for a total of 2,905 sentences. Similarly, for Nepali--English, the \emph{test} set is composed of 850 sentences originally in English, and 850 originally in Nepali. This %
yielded 1,462 sentence pairs originally in English and 1,462 originally in Nepali, for a total of 2,924 sentence pairs. The composition of the rest of the sets can be found in Table \ref{tab:set_stats}.

\begin{table}[t]
    \small
    \centering
   \begin{tabular}{p{20pt}@{\hspace{2em}}rrrrrr}
\toprule
 \multirow{2}{*}{\bf \shortstack[l]{orig\\ lang}} %
&  \multicolumn{2}{c}{\emph{dev}} &  \multicolumn{2}{c}{\emph{devtest}} & \multicolumn{2}{c}{\emph{test}} \\
\cmidrule(lr){2-3}
\cmidrule(lr){4-5}
\cmidrule(lr){6-7}
 &{uniq} & {tot} & {uniq} & {tot} & {uniq} & {tot}  \\
\midrule\\
\multicolumn{7}{l}{\bf \NeEn}\\
\En& 693 &   1,181 &800& 1,393 & 850 &  1,462 \\
\Ne & 825&  1,378 &800 &   1,442   & 850&  1,462 \\%\cmidrule{3-6}
    &{\bf 1,518}& {\bf 2,559}& {\bf 1,600}  & {\bf 2,835} & {\bf 1,700}& {\bf  2,924}\\
 \\
\multicolumn{7}{l}{\bf \SiEn}\\
 \En &1,123 &  1,913&800   & 1,395   & 850&  1,465 \\
\Si&  565 &   985 &800 &  1,371  & 850&  1,440 \\%\cmidrule{3-6}
 & {\bf 1,688}&{\bf 2,898}&{\bf 1600} & {\bf 2,766} & {\bf 1700}&{\bf 2,905}  \\\\
\bottomrule
\end{tabular}
    \caption{Number of unique sentences (\emph{uniq}) and total number of sentence pairs (\emph{tot}) per \flores{} test set grouped by their original languages. %
    }
    \label{tab:set_stats}
\end{table}

In Appendix Table~\ref{tab:topics}, we present the aggregate distribution of topics per sentence for the datasets in \NeEn and \SiEnns, which shows a diverse representation of topics ranging from General (e.g. documents about tires, shoes and insurance), History (e.g. documents about history of the radar, the Titanic, etc.) to Law  and Sports. This richness of topics increases the difficulty of the set, as it requires models that are rather domain-independent. The full list of documents and topics is also in  Appendix, Table~\ref{tab:wiki_docs}.

\section{Experiments} \label{sec:experiments}
In this section, we first describe the data used for training the models, we then discuss the learning settings and models considered, and finally we report the results of these baseline models on the new evaluation benchmarks. 

\subsection{Training Data} \label{subsec:trainingdata}
Small amounts of parallel data are available for Sinhala--English and Nepali--English. Statistics can be found in Table~\ref{tab:parallel-data}.
This data comes from different sources. Open Subtitles and GNOME/KDE/Ubuntu come from the OPUS repository\footnote{\url{http://opus.nlpl.eu/}}.
Global Voices is an updated version (2018q4) of a data set originally created for the CASMACAT project\footnote{\url{ http://casmacat.eu/corpus/global-voices.html}}.
Bible translations come from the bible-corpus\footnote{\url{https://github.com/christos-c/bible-corpus/}}.
The Paracrawl corpus comes from the Paracrawl project\footnote{\url{https://paracrawl.eu/}}.
The filtered version (Clean Paracrawl) was generated using the LASER model \cite{artetxe2018multi} to get the best sentence pairs having 1 million English tokens as specified in \citet{chaudhary-EtAl:2019:WMT}.
We also contrast this filtered version with a randomly filtered version (Random Paracrawl) with the same number of English tokens.
Finally, our multi-lingual experiments in Nepali use Hindi monolingual (about 5 million sentences) and English-Hindi parallel data (about 1.5 million parallel sentences) from the IIT Bombay corpus\footnote{\url{http://www.cfilt.iitb.ac.in/iitb_parallel/}}.

\begin{table}[t!]
\small
    \centering
\begin{tabular}{lrr}
    \toprule
  & {\bf Sentences} & {\bf Tokens} \\
  
   \multicolumn{3}{l}{\bf \NeEn}\\\midrule
\emph{parallel}\\
~ Bible                &  62K & 1.5M\\
~ Global Voices        &   3K &    75K\\
~ Penn Tree Bank       &   4K &    88K\\
~ GNOME/KDE/Ubuntu     & 495K & 2M\vspace{1ex}\\
\emph{comparable$^*$}\\
~ Unfiltered Paracrawl & 2.2M & 40.6M\\
~ Clean Paracrawl      & 32.9K & 1M\\
~ Random Paracrawl     & 55.3K & 1M\vspace{1ex}\\
\emph{monolingual}\\
~ Wikipedia (en)    &   67.8M &     2.0B\\
~ Common Crawl (ne) &    3.6M &   103.0M\\
~ Wikipedia (ne)    &   92.3K &     2.8M\\\\
\midrule
    \multicolumn{3}{l}{\bf \SiEn }\\\midrule
\emph{parallel} \\
~ Open Subtitles      & 601K & 3.6M\\
~ GNOME/KDE/Ubuntu     &  46K &   151K\vspace{1ex}\\
\emph{comparable$^*$}\\
~ Paracrawl & 3.4M & 45.4M \\
~ Clean Paracrawl      & 47K & 1M\\
~ Random Paracrawl     & 74.2K & 1M\vspace{1ex}\\ 
\emph{monolingual}\\
~ Wikipedia (en)    &   67.8M &     2.0B\\
~ Common Crawl (si) &    5.2M &   110.3M\\
~ Wikipedia (si)    &  155.9K &     4.7M\\

\bottomrule
    \end{tabular}
    \caption{Parallel, comparable, and monolingual data used in experiments in \textsection\ref{sec:experiments}. The number of tokens for parallel and comparable corpora are reported over the English tokens.
    Monolingual and comparable corpora do not include any sentences from the evaluation sets.
    \\
    $^*$Comparable data from Paracrawl is used only in the weakly-supervised experiments since alignments are noisy.
}
    \label{tab:parallel-data}
    \vspace{-15pt}
\end{table}

\subsection{Training Settings} \label{sec:settings}
We evaluate models in four training settings.
First, we consider a fully {\em supervised} training setting using the parallel data listed in Table~\ref{tab:parallel-data}.

Second, we consider a fully {\em unsupervised} setting, whereby only monolingual data on both the source and target side are used to train the model~\cite{lample_emnlp2018}.

Third, we consider a {\em semi-supervised} setting where %
we also leverage monolingual data on the target side using
the standard back-translation training protocol~%
\cite{sennrich2015improving}: %
we train a backward MT system, which we use to translate monolingual target sentences to the source language.
Then, we merge the resulting pairs of noisy (back-translated) source sentences with the original target sentences and add them as additional parallel data for training source-to-target MT system. Since monolingual data is available for both languages, we train backward MT systems in both directions and repeat the back-translation process iteratively~\cite{he2016dual, unsupNMTlample}.
We consider up to two back-translation iterations.
At each iteration we generate back-translations using beam search, which has been shown to perform well in low-resource settings~\cite{edunov2018backtranslation}; we use a beam width of 5 and individually tune the length-penalty on the \emph{dev} set.

Finally, we consider a {\em weakly supervised} setting by using a baseline system to filter out Paracrawl data using LASER ~\cite{artetxe2018multi} by following the approach similar to  \citet{chaudhary-EtAl:2019:WMT}, in order to augment the original training set with a possibly larger but noisier set of parallel sentences.

For Nepali only, we also consider training using Hindi data, both in a joint supervised and semi-supervised setting. For instance, at each iteration of the joint semi-supervised setting, we use models from the previous iteration to back-translate English monolingual data into both Hindi and Nepali, and from Hindi and Nepali monolingual data into English. We then concatenate actual parallel data and back-translated data of the same language pair together, and train a new model. We also consider using English-Hindi data in the unsupervised scenario. In that setting, a model is pretrained in an unsupervised way with English, Hindi and Nepali monolingual data using the unsupervised approach by \citet{lample2019cross}, and it is then
jointly trained on both the \NeEn unsupervised learning task and the Hindi-English supervised task (in both directions).

\iffalse
\begin{table*}[t]
\centering
\begin{tabular}[b]{lcccccc}
\toprule
 & \multicolumn{2}{c}{{\bf Supervised}} & \multicolumn{1}{c}{{\bf Weakly Supervised}}  & \multicolumn{2}{c}{{\bf Semi-supervised}} & \multicolumn{1}{c}{{\bf Unsupervised}} \\
 \cmidrule(lr){2-3}
 \cmidrule(lr){4-4}
 \cmidrule(lr){5-6}
 \cmidrule(lr){7-4}
 & PBSMT  & NMT & NMT & NMT (\#1) & NMT (\#2) & NMT \\
\midrule
{\bf \RNeEn}& \bleu{2.13} & 
\bleu{4.27} & - & \bleu{6.83} & \bleu{6.84} & \bleu{0.0} \\
{\bf \NeEn} & %
\bleu{3.5} & \bleu{7.57} & \bleu{7.5} & \bleu{12.725} & \bleu{15.061} & \bleu{0.07} \\
\midrule
{\bf \RSiEn} &
\bleu{1.38} & \bleu{1.23} & - & \bleu{5.22} & \bleu{6.46} & \bleu{0.08} \\
{\bf \SiEn} &
\bleu{4.8} & \bleu{7.159} & \bleu{9.9} & \bleu{12.046} & \bleu{15.105} & \bleu{0.08} \\
\bottomrule
\end{tabular}
\caption{BLEU scores of various machine translation methods and learning settings on \emph{devtest} (see \textsection\ref{sec:methodology}).
We report detokenized SacreBLEU~\cite{post2018sacrebleu} for \{Ne,Si\}$\rightarrow$En and tokenized BLEU for En$\rightarrow$\{Ne,Si\}.
}
\label{tab:bigtable}
\end{table*}
\fi

\begin{table*}[t]
\small
\centering
\begin{tabular}[b]{lccccccccc}
\toprule
 & \multicolumn{2}{c}{{\bf Supervised}} & \multicolumn{2}{c}{{\bf Unsupervised}}  & \multicolumn{4}{c}{{\bf Semi-supervised}} & \multicolumn{1}{c}{{\bf Weakly supervised}} \\

\cmidrule(lr){2-3}
\cmidrule(lr){4-5}
\cmidrule(lr){6-9}

& & +mult. & & + mult. & it. 1  & it. 2 & it 1. + mult. & it 2. + mult. &  \\
\midrule

{\bf \RNeEn} & \bleu{4.27} & \bleu{6.86} &  \bleu{0.11} & \bleu{8.34} & \bleu{6.83}
& \bleu{6.84} & \bleu{8.76} & \bleu{8.77} & \bleu{5.82} \\
{\bf \NeEn} & \bleu{7.57} & \bleu{14.18} & \bleu{0.47} & \bleu{18.78} &  \bleu{12.725}&  \bleu{15.061} & \bleu{19.78} &  \bleu{21.45} & \bleu{9.6} \\
\midrule
{\bf \RSiEn} & \bleu{1.23} & - & \bleu{0.08} & - & \bleu{5.22} & \bleu{6.46} & -  & - & \bleu{3.10} \\
{\bf \SiEn} & \bleu{7.159} & - & \bleu{0.08} & -  & \bleu{12.046} &  \bleu{15.105} & - & - & \bleu{10.86}\\
\bottomrule
\end{tabular}
\caption{BLEU scores of NMT using various  learning settings on \emph{devtest} (see \textsection\ref{sec:methodology}).
We report detokenized SacreBLEU~\cite{post2018sacrebleu} for \{Ne,Si\}$\rightarrow$En and tokenized BLEU for En$\rightarrow$\{Ne,Si\}.
}
\label{tab:bigtable}
\vspace{-3pt}
\end{table*}

\iffalse %
\begin{table*}[t]
\centering
\begin{tabular}[b]{lcccccc}
\toprule
 & \multicolumn{2}{c}{{\bf Supervised}} & \multicolumn{2}{c}{{\bf Semi-supervised}} & \multicolumn{2}{c}{{\bf Unsupervised}} \\
 \cmidrule(lr){2-3}
 \cmidrule(lr){4-5}
 \cmidrule(lr){6-7}
 & PBSMT  & NMT & NMT (iter.~1) & NMT (iter.~2) & PBSMT & NMT \\
\midrule
{\bf \RNeEn}& - & 
\bleu{4.27} & \bleu{6.83} & \bleu{6.84} & \bleu{0.0} & \bleu{0.0} \\
{\bf \NeEn} & %
\bleu{3.5} & \bleu{7.57} & \bleu{12.725} & \bleu{15.061} & \bleu{0.1} & \bleu{0.07} \\
\midrule
{\bf \RSiEn} &
- & \bleu{1.23} & \bleu{5.22} & \bleu{5.21} & \bleu{0.0} & \bleu{0.08} \\
{\bf \SiEn} &
\bleu{4.8} & \bleu{7.159} & \bleu{12.046} & \bleu{} & \bleu{0.1} & \bleu{0.08} \\
\bottomrule
\end{tabular}
\caption{BLEU scores of various machine translation methods and learning settings on \emph{devtest} (see \textsection\ref{sec:methodology}).
We report detokenized SacreBLEU~\cite{post2018sacrebleu} for \{Ne,Si\}$\rightarrow$En and tokenized BLEU for En$\rightarrow$\{Ne,Si\}.
}
\label{tab:bigtable}
\end{table*}
\fi

\subsection{Models \& Architectures}
We consider both phrase-based statistical machine translation (PBSMT) and neural machine translation (NMT) systems in our experiments. All hyper-parameters have been cross-validated using the dev set.
The PBSMT systems use Moses~\cite{moses}, with
state-of-the-art settings (5-gram language model, hierarchical lexicalized reordering model, operation sequence model) but no additional monolingual data to train the language model.

The NMT systems use the Transformer~\citep{vaswani2017attention}
implementation in the Fairseq toolkit~\citep{ott2019fairseq}; preliminary experiments showed these to perform better than LSTM-based NMT models.
More specifically, in the supervised setting,
we use a Transformer architecture with 5 encoder and 5 decoder layers, where the number of attention heads, embedding dimension and inner-layer dimension are 2, 512 and 2048, respectively.
In the semi-supervised setting, where we augment our small parallel training data with millions of back-translated sentence pairs, we use a larger Transformer architecture with 6 encoder and 6 decoder layers, where the number of attention heads, embedding dimension and inner-layer dimension are 8, 512 and 4096, respectively.
When we use multilingual data, the encoder is shared in the \HiNeEn direction, and the decoder is shared in the \RHiNeEn  direction. 
We regularize our models with dropout, label smoothing and weight decay, with the corresponding hyper-parameters tuned independently for each language pair.
Models are optimized with Adam~\citep{kingma:adam:2015} using $\beta_1 = 0.9$, $\beta_2 = 0.98$, and $\epsilon = 1e-8$.
We use the same learning rate schedule as \citet{ott:scaling:2018}.
We run experiments on between 4 and 8 Nvidia V100 GPUs with mini-batches of between 10K and 100K target tokens following~\citet{ott:scaling:2018}.
Code to reproduce our results can be found at {\small \url{https://github.com/facebookresearch/flores}}.

\subsection{Preprocessing and Evaluation}

We tokenize Nepali and Sinhala using the Indic NLP Library.\footnote{\url{https://github.com/anoopkunchukuttan/indic_nlp_library}}
For the PBSMT system, we tokenize English sentences using the Moses tokenization scripts.
For NMT systems, we instead use a vocabulary of 5K symbols based on a joint source and target Byte-Pair Encoding (BPE;~\citealp{sennrich2015improving}) learned using the sentencepiece library\footnote{\url{https://github.com/google/sentencepiece}} over the parallel training data.
We learn the joint BPE for each language pair over the raw English sentences and tokenized Nepali or Sinhala sentences.
We then remove training sentence pairs with more than 250 source or target BPE tokens.

We report detokenized SacreBLEU~\cite{post2018sacrebleu} when translating into English, and tokenized BLEU~\cite{bleu} when translating from English into Nepali or Sinhala.

\subsection{Results}

In the supervised setting, PBSMT performed quite worse than NMT, achieving BLEU scores of 2.5, 4.4, 1.6 and 5.0 on \RNeEnns, \NeEnns, \RSiEn and \SiEnns, respectively. 
Table~\ref{tab:bigtable} reports results using NMT in all the other learning configurations described in  \textsection\ref{sec:settings}. There are several observations we can make.

First, these language pairs are very difficult, as even supervised NMT baselines achieve BLEU scores less than 8.
Second and not surprisingly, the BLEU score is particularly low when translating into the more morphologically rich Nepali and Sinhala languages.
Third, unsupervised NMT approaches seem to be ineffective on these distant language pairs, achieving BLEU scores close to 0. 
The reason for this failure is due to poor initialization of the word embeddings. %

Poor initialization can be attributed to
to the %
monolingual corpora used to train word embeddings which do not have sufficient number of overlapping strings, and are not comparable~\cite{neubig18emnlp, sogaard18}.

Fourth, the biggest improvements are brought by the semi-supervised approach using back-translation, which nearly doubles BLEU for \NeEn from 7.6 to 15.1 (+7.5 BLEU points) and \SiEn from 7.2 to 15.1 (+7.9 BLEU points), and increases +2.5 BLEU points for \RNeEn and +5.3 BLEU points for \RSiEnns.

Fifth, additional parallel data in English-Hindi further improves translation quality in Nepali across all settings. For instance, in the \NeEn supervised setting, we observe a gain of 6.5 BLEU points, while in the semi-supervised setting (where we back-translate also to and from Hindi) the gain is 6.4 BLEU points. %
Similarly, in the unsupervised setting, multilingual training with Hindi brings \NeEn to 3.9 BLEU and \RNeEn to 2.5 BLEU; if however, the architecture is pretrained as prescribed by~\citet{lample2019cross}, BLEU score improves %
to 18.8 BLEU for \NeEn and 8.3 BLEU for \RNeEnns.

Finally, the weakly supervised baseline using the additional noisy parallel data described in \textsection\ref{subsec:trainingdata} improves upon the supervised baseline in all four directions. This is studied in more depth in Table~\ref{tab:corpusablation2} for \SiEn and \NeEnns. Without any filtering or with random filtering, BLEU score is close to 0 BLEU. Applying the a filtering method based on LASER scores \cite{artetxe2018multi} provides an improvement over using the unfiltered Paracrawl, of +5.5 BLEU points for \NeEn and +7.3 BLEU points for \SiEnns. Adding Paracrawl Clean to the initial parallel data improves performance by +2.0 and +3.7 BLEU points, for \NeEn and \SiEnns, respectively.

\begin{table}[hbt]
\small
\begin{center}
\begin{tabular}{lrrrr}
\toprule
\multirow{2}{*}{\bf Corpora} & \multicolumn{2}{c}{\bf BLEU} \\
\cmidrule{2-3}
 & \neen & %
 \sien\\
\midrule
Parallel & %
\bleu{7.6} & %
\bleu{7.2}  \\
Unfiltered Paracrawl & %
\bleu{0.43} & %
\bleu{0.44} \\
Paracrawl Random & %
\bleu{0.14} & %
\bleu{0.36}\\
Paracrawl Clean & %
\bleu{5.9} & %
\bleu{7.73}\\
Parallel + Paracrawl Clean & %
\bleu{9.60} & %
\bleu{10.86}\\
\bottomrule
\end{tabular}
\end{center}
\caption{\small Weakly supervised experiments: Adding noisy parallel data from filtered Paracrawl improves translation quality in some conditions. ``Parallel'' refers to the data described in Table~\ref{tab:parallel-data}.} 
\label{tab:corpusablation2}
\end{table}

\section{Discussion}
\begin{figure*}[!t]
\begin{center}
\includegraphics[scale=0.4]{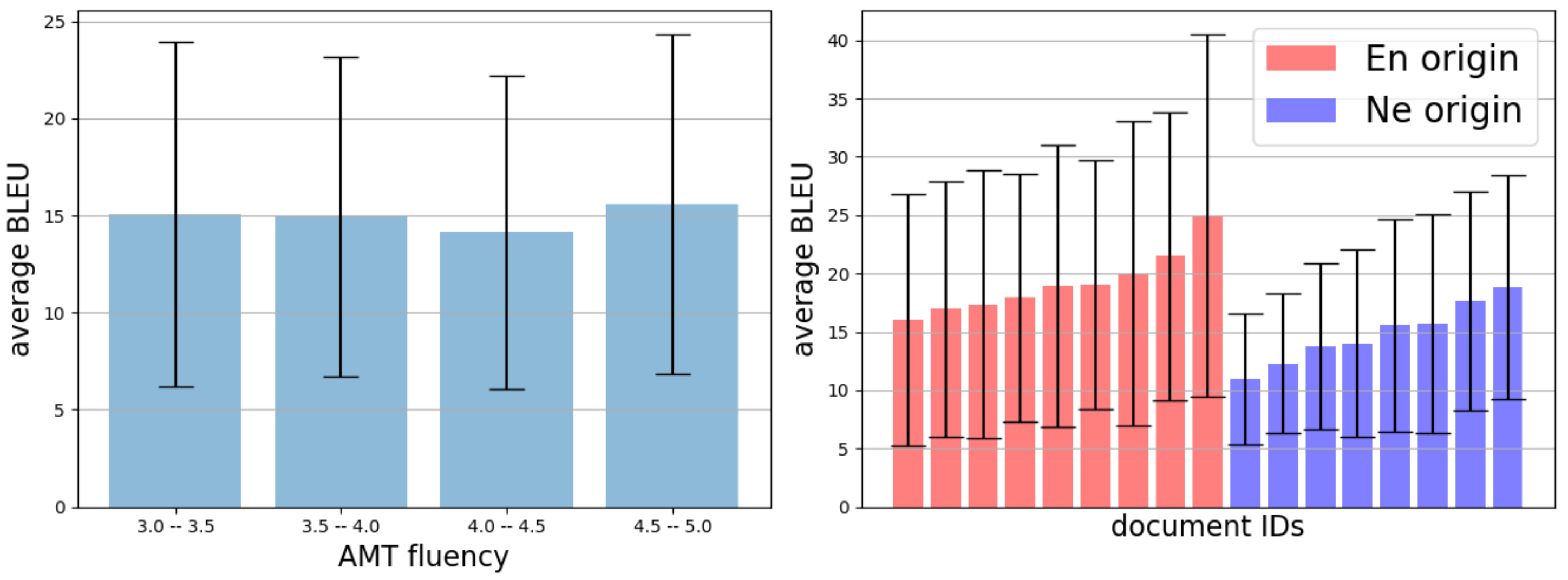}
\end{center}
\vspace{0.1cm}
\caption{\small Analysis of the Ne$\rightarrow$En \emph{devtest} set using the semi-supervised machine translation system.
{\bf Left}: sentence level BLEU versus AMT fluency score of the reference sentences in English; source sentences that have received more fluent human translations are not easier to translate by machines.
{\bf Right}: average sentence level BLEU against Wikipedia document id from which the source sentence was extracted; sentences have roughly the same degree of difficulty across documents since there is no extreme difference between shortest and tallest bar. However, source sentences originating from Nepali Wikipedia (blue) are translated more poorly than those originating from English Wikipedia (red). Documents are sorted by BLEU for ease of reading.}
\label{fig:analysis}
\end{figure*}

In this section, we provide an analysis of the performance on the Nepali to English \emph{devtest} set using the semi-supervised machine translation system, see Figure~\ref{fig:analysis}. Findings on other language directions are similar.

\noindent
{\bf Fluency of references:} we observe no correlation between the fluency rating of human references and the quality of translations as measured by BLEU. This suggests that %
the difficulty of the translation task is not related to the fluency of the references, %
at least at the current level of accuracy. 

\noindent
{\bf Document difficulty:} we observe that translation quality is similar %
across all document ids, with a difference of 10 BLEU points between the document that is the easiest and the hardest to translate. This suggests that the random sampling procedure used to construct the dataset was adequate and that no single Wikipedia document produces much harder sentences than others.

\noindent
{\bf Original vs translationese:} %
we noticed that documents originating from Nepali are harder to translate than documents originating in English. This holds when performing the evaluation with the supervised MT system: translations of original Nepali sentences obtain 4.9 BLEU while Nepali translationese %
obtain 9.1 BLEU. This suggests that the existing parallel corpus is closer to English Wikipedia than Nepali Wikipedia.

\subsection{Domain drift} 
To better understand the effect of domain mismatch between the parallel dataset and the Wikipedia evaluation set, we restricted the \SiEn training set to only the Open Subtitles portion of the parallel dataset, and we held out 1000 sentences for ``in-domain'' evaluation of generalization performance. 
Table~\ref{tab:domain} shows that translation quality on in-domain data is between 10 and 16 BLEU points higher.
This may be due to both domain mismatch as well as sensitivity of the BLEU metric to sentence length.
Indeed, there are on average 6 words per sentences in the Open Subtitles test set compared to 16 words per sentence in the \flores{} \emph{devtest} set.
However, when we train semi-supervised models on back-translated Wikipedia data whose domain better matches the ``Out-of-domain'' \emph{devtest} set, we see much larger gains in BLEU for the ``Out-of-domain'' set than we see on the ``In-domain'' set, suggesting that domain mismatch is indeed a major problem.

\begin{table}[hbt]
\small
\centering
\begin{tabular}{lll}
\toprule 
& {\bf Open Subtitles} & \bf \flores{} (\emph{devtest}) \\
\midrule
\multicolumn{3}{l}{\bf \SiEn} \\
\quad Supervised & \enskip \bleu{23.457} & \enskip \bleu{7.159} \\
\quad Semi-sup. & \enskip \bleu{28.148} (+20\%) & \enskip \bleu{15.105} (+210\%) \\
\midrule
\multicolumn{3}{l}{\bf \RSiEn} \\
\quad Supervised & \enskip \bleu{10.99} & \enskip \bleu{1.23} \\
\quad Semi-sup. & \enskip \bleu{11.76} (+7\%) & \enskip \bleu{6.46} (+542\%) \\
\bottomrule
\end{tabular}
\caption{In-domain vs.~out-of-domain translation performance (BLEU) for supervised and semi-supervised NMT models.
We report BLEU on a held-out subset of 1,000 sentences from the Open Subtitles training data (see Table~\ref{tab:parallel-data}) and on \emph{devtest} (see \textsection\ref{sec:methodology}).
Semi-supervised models are trained on back-translated Wikipedia data.
\vspace{-3pt} }
\label{tab:domain}
\end{table}
\section{Conclusions}
One of the biggest challenges in MT today is learning to translate low-resource language pairs. Research in this area not only faces formidable technical challenges, from  learning with limited supervision to dealing with very distant languages, but it is also hindered by the lack of freely and publicly available evaluation benchmarks. 

In this work, we introduce and freely release to the community \flores{} benchmarks for \NeEn and \SiEn. Nepali and Sinhala are languages with very different syntax and morphology than English; also, very little parallel data in these language pairs is publicly available. However, a good amount of monolingual data, parallel data in related languages, and Paracrawl data exist in both languages, making these two language pairs a perfect candidate for research on low-resource MT.

Our experiments show that current state-of-the-art approaches perform rather poorly on these new evaluation benchmarks, with semi-supervised and in particular multi-lingual neural methods outperforming all the other model variants and training settings we considered. We perform additional analysis to probe the quality of the datasets. We find no evidence of poor construction quality, yet observe that the low BLEU scores are partly due to the domain mismatch between the training and test datasets.  %
We believe that these benchmarks will help the research community on low-resource MT make faster progress by enabling free access to evaluation data on actual low-resource languages and promoting fair comparison of methods. 

\bibliography{paper}
\bibliographystyle{acl_natbib}

\newpage
\clearpage
\appendix
\onecolumn
\section{List of Topics}
\begin{table}[!h]
   \small
    \centering
    \begin{tabular}{p{3cm}rr}
    \toprule
    \centering
    \multirow{2}{*}{\bf topic}& \multicolumn{2}{c}{\bf proportion (\%)}\\\cmidrule{2-3}
    	&{\neen}&{\sien}\\
    \midrule
    General	&18.3&	24.1\\
    History	&6.5&	15.1\\
    Science	&7.4&	12.7\\
    Religion&	8.9&	10.5\\
    Social Sciences&	10.2&	6.9\\
    Biology&	6.3&	9.1 \\
    Geography&	10.6&	4.6\\
    Art/Culture&	6.7&	8.3\\
    Sports&	5.8	&6.7\\
    Politics&	8.1&	N/A\\
    People&	7.4	&N/A\\
    Law	&3.9& 2.0\\
    \bottomrule
    \end{tabular}
    \caption{Distribution of the topics of the sentences in the dev, devtest and test sets according to the  Wikipedia document they were sampled from.}
    \label{tab:topics}
\end{table}

\clearpage
\section{Statistics of automatic filtering and manual filtering}

\begin{figure}[!h]
    \centering
    \begin{minipage}{0.5\textwidth}
        \centering
        \includegraphics[width=1.0\textwidth]{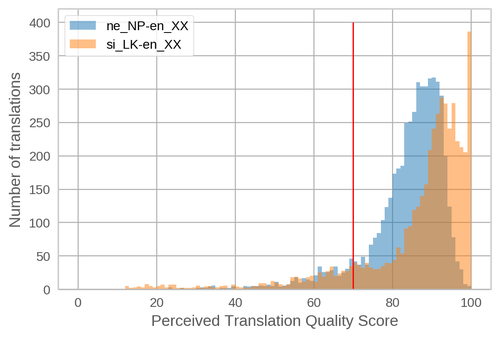} %
    \end{minipage}\hfill
    \begin{minipage}{0.5\textwidth}
        \centering
        \includegraphics[width=1.0\textwidth]{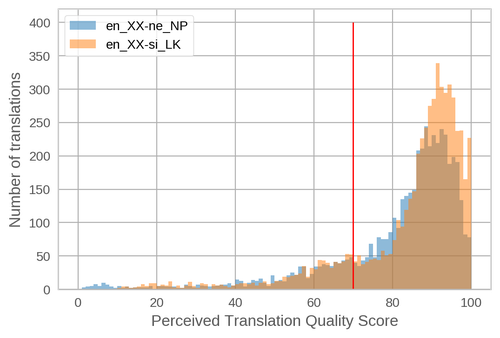} %
    \end{minipage}
    \caption{Histogram of averaged translation quality score. We ask three different raters to rate each sentence from 0--100 according to the perceived translation quality. In our guidelines, the 0--10 range represents a translation that is completely incorrect and inaccurate; the 11--29 range represents a translation with few correct keywords, but the overall meaning is different from the source; the 30--50 range represents a translation that contains translated fragments of the source string, with major mistakes; the 51--69 range represents a translation which is understandable and conveys the overall meaning of source string but contains typos or grammatical errors; the 70--90 range represents a translation 
that closely preserves the semantics of the source sentence;  and the 90--100 range represents a \emph{perfect} translation. Translations with averaged translation score less than 70 (red line) are removed from the dataset.}
\end{figure}

\begin{figure}[!h]
    \centering
    \begin{minipage}{0.5\textwidth}
        \centering
        \includegraphics[width=1.0\textwidth]{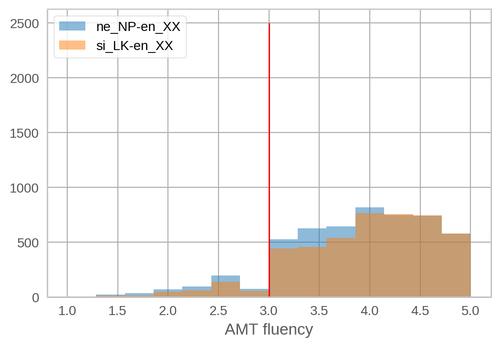} %
    \end{minipage}\hfill
    \caption{Histogram of averaged AMT fluency score of English translations. We ask five different raters to rate each sentence from 1--5 according to its fluency. In our guidelines, the 1--2 range represents a sentence that is not fluent, 3 is neutral, while the 4--5 range is for fluent sentences that raters can easily understand. Translations with averaged fluency score less than 3 (red line) are removed from the dataset.}
\end{figure}

\begin{table*}[!h]
    \centering
    \scriptsize
    \begin{tabular}{lllll}
    \toprule
     & {\bf \NeEn} & {\bf \RNeEn} & {\bf \SiEn} & {\bf \RSiEn} \\
    \midrule
        {\bf Automatic filtering} & 14\% & 18\% & 24\% & 7\% \\
        {\bf Manual filtering} \\
        ~ Translation quality & 10\% & 19\% & 13\% & 16\% \\
        ~ Fluency & 10\% & - & 17\% & - \\
    \bottomrule

    \end{tabular}
    \caption{Percentage of translations that did not pass the automatic and manual filtering checks. We first use automatic methods to filter out poor translations and send those translations back for rework. We then collect translations that pass the automatic filtering and send them to two human quality checks, one for adequacy and the other for fluency. Note that the percentage of sentences that did not pass manual filtering is among those sentences that passed the automatic filtering.}
    \label{tab:ratio_filtered}
\end{table*}

\clearpage
\section{List of Wikpedia Documents}
\begin{table*}[!h]
    \centering
    \scriptsize
    \begin{tabular}{lll}
    \toprule
    {\bf domain} & {\bf document/gloss} & {\bf topic} \\
    \midrule
en.wikipedia.org&\href{http://en.wikipedia.org/wiki/Astronomy}{\texttt{Astronomy}} & Science \\
en.wikipedia.org&\href{http://en.wikipedia.org/wiki/History_of_radar}{\texttt{History of radar}} & History \\
en.wikipedia.org&\href{http://en.wikipedia.org/wiki/Shoe}{\texttt{Shoe}} & General \\
en.wikipedia.org&\href{http://en.wikipedia.org/wiki/Tire}{\texttt{Tire}} & General \\
en.wikipedia.org&\href{http://en.wikipedia.org/wiki/Indian_cuisine}{\texttt{Indian cuisine}} & Art/Culture \\
en.wikipedia.org&\href{http://en.wikipedia.org/wiki/IPhone}{\texttt{IPhone}} & General \\
en.wikipedia.org&\href{http://en.wikipedia.org/wiki/Apollo_program}{\texttt{Apollo program}} & History \\
en.wikipedia.org&\href{http://en.wikipedia.org/wiki/Chess}{\texttt{Chess}} & General \\
en.wikipedia.org&\href{http://en.wikipedia.org/wiki/Honey}{\texttt{Honey}} & General \\
en.wikipedia.org&\href{http://en.wikipedia.org/wiki/Police}{\texttt{Police}} & Law \\
en.wikipedia.org&\href{http://en.wikipedia.org/wiki/Desert}{\texttt{Desert}} & Geography \\
en.wikipedia.org&\href{http://en.wikipedia.org/wiki/Slavery}{\texttt{Slavery}} & Social Sciences \\
en.wikipedia.org&\href{http://en.wikipedia.org/wiki/Riddler}{\texttt{Riddler}} & Art/Culture \\
en.wikipedia.org&\href{http://en.wikipedia.org/wiki/Diving}{\texttt{Diving}} & Sports \\
en.wikipedia.org&\href{http://en.wikipedia.org/wiki/Cat}{\texttt{Cat}} & Biology \\
en.wikipedia.org&\href{http://en.wikipedia.org/wiki/Boxing}{\texttt{Boxing}} & Sports \\
en.wikipedia.org&\href{http://en.wikipedia.org/wiki/White_wine}{\texttt{White wine}} & General \\
en.wikipedia.org&\href{http://en.wikipedia.org/wiki/Creativity}{\texttt{Creativity}} & Social Sciences \\
en.wikipedia.org&\href{http://en.wikipedia.org/wiki/Capitalism}{\texttt{Capitalism}} & Social Sciences \\
en.wikipedia.org&\href{http://en.wikipedia.org/wiki/Alaska}{\texttt{Alaska}} & Geography \\
en.wikipedia.org&\href{http://en.wikipedia.org/wiki/Museum}{\texttt{Museum}} & General \\
en.wikipedia.org&\href{http://en.wikipedia.org/wiki/Lifeguard}{\texttt{Lifeguard}} & General \\
en.wikipedia.org&\href{http://en.wikipedia.org/wiki/Tennis}{\texttt{Tennis}} & Sports \\
en.wikipedia.org&\href{http://en.wikipedia.org/wiki/Writer}{\texttt{Writer}} & General \\
en.wikipedia.org&\href{http://en.wikipedia.org/wiki/Anatomy}{\texttt{Anatomy}} & Science \\
si.wikipedia.org&\href{http://si.wikipedia.org/wiki/\%E0\%B6\%9A\%E0\%B7\%94\%E0\%B6\%BB\%E0\%B7\%8A\%E0\%B6\%86\%E0\%B6\%B1\%E0\%B7\%8A}{\texttt{Qoran}} & Religion \\
si.wikipedia.org&\href{http://si.wikipedia.org/wiki/\%E0\%B6\%B0\%E0\%B6\%B8\%E0\%B7\%8A\%E2\%80\%8C\%E0\%B6\%B8\%E0\%B7\%84\%E0\%B6\%AF\%E0\%B6\%BA\%E0\%B7\%80\%E0\%B7\%92\%E0\%B6\%B7\%E0\%B6\%9E\%E0\%B7\%8A\%E2\%80\%8C\%E0\%B6\%9C\%E0\%B7\%9C}{\texttt{Dhammas}} & Religion \\
si.wikipedia.org&\href{http://si.wikipedia.org/wiki/\%E0\%B7\%83\%E0\%B7\%8A\%E0\%B7\%80\%E0\%B6\%B7\%E0\%B7\%8F\%E0\%B7\%80\%E0\%B7\%92\%E0\%B6\%9A_\%E0\%B7\%80\%E0\%B7\%98\%E0\%B6\%9A\%E0\%B7\%8A\%E2\%80\%8D\%E0\%B7\%82\%E0\%B6\%BD\%E0\%B6\%AD\%E0\%B7\%8F}{\texttt{Vegetation}} & Science \\
si.wikipedia.org&\href{http://si.wikipedia.org/wiki/\%E0\%B6\%9A\%E0\%B7\%9C\%E0\%B7\%85\%E0\%B6\%B9_\%E0\%B6\%B1\%E0\%B7\%8F\%E0\%B6\%BD\%E0\%B6\%B1\%E0\%B7\%8A\%E0\%B6\%AF_\%E0\%B7\%80\%E0\%B7\%92\%E0\%B6\%AF\%E0\%B7\%94\%E0\%B7\%84\%E0\%B6\%BD\%E0\%B7\%9A_\%E0\%B6\%A0\%E0\%B7\%92\%E0\%B6\%BB\%E0\%B6\%B4\%E0\%B7\%8A\%E2\%80\%8D\%E0\%B6\%BB\%E0\%B6\%9A\%E0\%B6\%A7_\%E0\%B6\%86\%E0\%B6\%AF\%E0\%B7\%92_\%E0\%B7\%83\%E0\%B7\%92\%E0\%B7\%83\%E0\%B7\%94_\%E0\%B6\%BD\%E0\%B7\%90\%E0\%B6\%BA\%E0\%B7\%92\%E0\%B7\%83\%E0\%B7\%8A\%E0\%B6\%AD\%E0\%B7\%94\%E0\%B7\%80}{\texttt{Names of Colombo Students}} & History \\
si.wikipedia.org&\href{http://si.wikipedia.org/wiki/\%E0\%B6\%A7\%E0\%B6\%BA\%E0\%B7\%92\%E0\%B6\%A7\%E0\%B7\%90\%E0\%B6\%B1\%E0\%B7\%92\%E0\%B6\%9A\%E0\%B7\%8A_\%28\%E0\%B6\%B1\%E0\%B7\%9E\%E0\%B6\%9A\%E0\%B7\%8F\%E0\%B7\%80\%29}{\texttt{Titanic}} & History \\
si.wikipedia.org&\href{http://si.wikipedia.org/wiki/\%E0\%B7\%84\%E0\%B7\%98\%E0\%B6\%AF\%E0\%B6\%BA}{\texttt{The Heart}} & Biology \\
si.wikipedia.org&\href{http://si.wikipedia.org/wiki/\%E0\%B6\%9A\%E0\%B6\%B1}{\texttt{The Ear}} & Biology \\
si.wikipedia.org&\href{http://si.wikipedia.org/wiki/\%E0\%B6\%AE\%E0\%B7\%9A\%E0\%B6\%BB\%E0\%B7\%80\%E0\%B7\%8F\%E0\%B6\%AF\%E0\%B6\%BA}{\texttt{Theravada}} & Religion \\
si.wikipedia.org&\href{http://si.wikipedia.org/wiki/\%E0\%B7\%80\%E0\%B7\%96_\%E0\%B7\%83\%E0\%B7\%9A\%E0\%B6\%A7\%E0\%B7\%92\%E0\%B6\%BA\%E0\%B7\%91\%E0\%B6\%B1\%E0\%B7\%8A}{\texttt{WuZetian}} & History \\
si.wikipedia.org&\href{http://si.wikipedia.org/wiki/\%E0\%B6\%B8\%E0\%B6\%B1\%E0\%B7\%9D_\%E0\%B7\%80\%E0\%B7\%92\%E0\%B7\%81\%E0\%B7\%8A\%E0\%B6\%BD\%E0\%B7\%9A\%E0\%B7\%82\%E0\%B6\%AB\%E0\%B7\%80\%E0\%B7\%8F\%E0\%B6\%AF\%E0\%B6\%BA}{\texttt{Psychoanalisis}} & Science \\
si.wikipedia.org&\href{http://si.wikipedia.org/wiki/\%E0\%B6\%85\%E0\%B6\%82\%E0\%B6\%9C\%E0\%B7\%94\%E0\%B6\%BD\%E0\%B7\%92\%E0\%B6\%B8\%E0\%B7\%8F\%E0\%B6\%BD_\%E0\%B6\%B4\%E0\%B7\%92\%E0\%B6\%BB\%E0\%B7\%92\%E0\%B6\%AD}{\texttt{Angulimala}} & Religion \\
si.wikipedia.org&\href{http://si.wikipedia.org/wiki/\%E0\%B6\%BB\%E0\%B6\%9A\%E0\%B7\%8A\%E2\%80\%8D\%E0\%B7\%82\%E0\%B6\%AB\%E0\%B6\%BA}{\texttt{Insurance}} & General \\
si.wikipedia.org&\href{http://si.wikipedia.org/wiki/\%E0\%B6\%B4\%E0\%B6\%AD\%E0\%B7\%8A\%E2\%80\%8D\%E0\%B6\%BB_\%E0\%B6\%9A\%E0\%B6\%BD\%E0\%B7\%8F\%E0\%B7\%80}{\texttt{Leafart}} & Art/Culture \\
si.wikipedia.org&\href{http://si.wikipedia.org/wiki/\%E0\%B6\%A2\%E0\%B6\%B1_\%E0\%B7\%83\%E0\%B6\%B1\%E0\%B7\%8A\%E0\%B6\%B1\%E0\%B7\%92\%E0\%B7\%80\%E0\%B7\%9A\%E0\%B6\%AF\%E0\%B6\%B1\%E0\%B6\%BA}{\texttt{Communication Science}} & Science \\
si.wikipedia.org&\href{http://si.wikipedia.org/wiki/\%E0\%B6\%B1\%E0\%B7\%99\%E0\%B7\%86\%E0\%B6\%BB\%E0\%B7\%8A\%E0\%B6\%B1\%E0\%B7\%99\%E0\%B7\%86\%E0\%B6\%BB\%E0\%B7\%94\%E0\%B6\%86\%E0\%B6\%A7\%E0\%B6\%B1\%E0\%B7\%8A}{\texttt{Pharaoh Neferneferuaten}} & History \\
ne.wikipedia.org&\href{http://ne.wikipedia.org/wiki/\%E0\%A4\%A8\%E0\%A5\%87\%E0\%A4\%B2\%E0\%A5\%8D\%E0\%A4\%B8\%E0\%A4\%A8_\%E0\%A4\%AE\%E0\%A4\%A8\%E0\%A5\%8D\%E0\%A4\%A1\%E0\%A5\%87\%E0\%A4\%B2\%E0\%A4\%BE}{\texttt{Nelson Mandela}} & People \\
ne.wikipedia.org&\href{http://ne.wikipedia.org/wiki/\%E0\%A4\%AD\%E0\%A4\%BE\%E0\%A4\%B0\%E0\%A4\%A4\%E0\%A5\%80\%E0\%A4\%AF_\%E0\%A4\%B8\%E0\%A4\%82\%E0\%A4\%B8\%E0\%A4\%A6}{\texttt{Parliament of India}} & Politics \\
ne.wikipedia.org&\href{http://ne.wikipedia.org/wiki/\%E0\%A4\%95\%E0\%A5\%88\%E0\%A4\%B2\%E0\%A4\%BE\%E0\%A4\%B2\%E0\%A5\%80_\%E0\%A4\%B0_\%E0\%A4\%95\%E0\%A4\%82\%E0\%A4\%9A\%E0\%A4\%A8\%E0\%A4\%AA\%E0\%A5\%81\%E0\%A4\%B0}{\texttt{Kailali and Kanchanpur}} & Geography \\
ne.wikipedia.org&\href{http://ne.wikipedia.org/wiki/\%E0\%A4\%AD\%E0\%A5\%81\%E0\%A4\%B5\%E0\%A4\%A8\%E0\%A4\%AA\%E0\%A5\%8B\%E0\%A4\%96\%E0\%A4\%B0\%E0\%A5\%80}{\texttt{Bhuwan Pokhari}} & Geography \\
ne.wikipedia.org&\href{http://ne.wikipedia.org/wiki/\%E0\%A4\%B8\%E0\%A5\%80\%E0\%A4\%93\%E0\%A4\%AA\%E0\%A5\%80\%E0\%A4\%A1\%E0\%A5\%80}{\texttt{COPD}} & Biology \\
ne.wikipedia.org&\href{http://ne.wikipedia.org/wiki/\%E0\%A4\%95\%E0\%A4\%BE\%E0\%A4\%B2\%E0\%A4\%B8\%E0\%A4\%B0\%E0\%A5\%8D\%E0\%A4\%AA_\%E0\%A4\%AF\%E0\%A5\%8B\%E0\%A4\%97}{\texttt{KaalSarp Yoga}} & Religion \\
ne.wikipedia.org&\href{http://ne.wikipedia.org/wiki/\%E0\%A4\%85\%E0\%A4\%B0\%E0\%A5\%8D\%E0\%A4\%A5\%E0\%A4\%B6\%E0\%A4\%BE\%E0\%A4\%B8\%E0\%A5\%8D\%E0\%A4\%A4\%E0\%A5\%8D\%E0\%A4\%B0\%E0\%A4\%AE\%E0\%A4\%BE_\%E0\%A4\%B6\%E0\%A5\%8B\%E0\%A4\%A7_\%E0\%A4\%85\%E0\%A4\%A8\%E0\%A5\%81\%E0\%A4\%B8\%E0\%A4\%A8\%E0\%A5\%8D\%E0\%A4\%A7\%E0\%A4\%BE\%E0\%A4\%A8}{\texttt{Research Methodology in Economics}} & Social Sciences \\
ne.wikipedia.org&\href{http://ne.wikipedia.org/wiki/\%E0\%A4\%A8\%E0\%A4\%BF\%E0\%A4\%AC\%E0\%A4\%A8\%E0\%A5\%8D\%E0\%A4\%A7}{\texttt{Essay}} & Social Sciences \\
ne.wikipedia.org&\href{http://ne.wikipedia.org/wiki/\%E0\%A4\%89\%E0\%A4\%A4\%E0\%A5\%8D\%E0\%A4\%AA\%E0\%A4\%B0\%E0\%A4\%BF\%E0\%A4\%B5\%E0\%A4\%B0\%E0\%A5\%8D\%E0\%A4\%A4\%E0\%A4\%A8}{\texttt{Mutation}} & Science \\
ne.wikipedia.org&\href{http://ne.wikipedia.org/wiki/\%E0\%A4\%8F_\%E0\%A4\%A8\%E0\%A5\%87\%E0\%A4\%95\%E0\%A4\%AA\%E0\%A4\%BE_\%E0\%A4\%AE\%E0\%A4\%BE\%E0\%A4\%93\%E0\%A4\%B5\%E0\%A4\%BE\%E0\%A4\%A6\%E0\%A5\%80\%E0\%A4\%95\%E0\%A4\%BE_\%E0\%A4\%B8\%E0\%A4\%82\%E0\%A4\%B5\%E0\%A4\%BF\%E0\%A4\%A7\%E0\%A4\%BE\%E0\%A4\%A8_\%E0\%A4\%B8\%E0\%A4\%AD\%E0\%A4\%BE\%E0\%A4\%B8\%E0\%A4\%A6}{\texttt{Maoist Constituent Assembly}} & Politics \\
ne.wikipedia.org&\href{http://ne.wikipedia.org/wiki/\%E0\%A4\%AA\%E0\%A4\%9F\%E0\%A4\%A8\%E0\%A4\%BE}{\texttt{Patna}} & Geography \\
ne.wikipedia.org&\href{http://ne.wikipedia.org/wiki/\%E0\%A4\%B8\%E0\%A4\%99\%E0\%A5\%8D\%E0\%A4\%98\%E0\%A5\%80\%E0\%A4\%AF_\%E0\%A4\%B6\%E0\%A4\%BE\%E0\%A4\%B8\%E0\%A4\%A8_\%E0\%A4\%AA\%E0\%A5\%8D\%E0\%A4\%B0\%E0\%A4\%A3\%E0\%A4\%BE\%E0\%A4\%B2\%E0\%A5\%80}{\texttt{Federal rule systen}} & Law \\
ne.wikipedia.org&\href{http://ne.wikipedia.org/wiki/\%E0\%A4\%A8\%E0\%A5\%87\%E0\%A4\%AA\%E0\%A4\%BE\%E0\%A4\%B2_\%E0\%A4\%B0_\%E0\%A4\%A8\%E0\%A5\%87\%E0\%A4\%B5\%E0\%A4\%BE\%E0\%A4\%B0_\%E0\%A4\%9C\%E0\%A4\%BE\%E0\%A4\%A4\%E0\%A4\%BF}{\texttt{Newari Community}} & Art/Culture \\
ne.wikipedia.org&\href{http://ne.wikipedia.org/wiki/\%E0\%A4\%B0\%E0\%A5\%88\%E0\%A4\%95\%E0\%A4\%BE_\%E0\%A4\%B0\%E0\%A4\%BE\%E0\%A4\%9C\%E0\%A4\%B5\%E0\%A4\%82\%E0\%A4\%B6}{\texttt{Raka's Dynasty}} & History \\
ne.wikipedia.org&\href{http://ne.wikipedia.org/wiki/\%E0\%A4\%A7\%E0\%A4\%BE\%E0\%A4\%A8}{\texttt{Rice}} & Biology \\
ne.wikipedia.org&\href{http://ne.wikipedia.org/wiki/\%E0\%A4\%B8\%E0\%A5\%8D\%E0\%A4\%A4\%E0\%A4\%A8\%E0\%A4\%AA\%E0\%A4\%BE\%E0\%A4\%A8}{\texttt{Breastfeeding}} & Biology \\
ne.wikipedia.org&\href{http://ne.wikipedia.org/wiki/\%E0\%A4\%AD\%E0\%A5\%81\%E0\%A4\%88\%E0\%A4\%82\%E0\%A4\%9A\%E0\%A4\%BE\%E0\%A4\%B2\%E0\%A5\%8B}{\texttt{Earthquake}} & Science \\
ne.wikipedia.org&\href{http://ne.wikipedia.org/wiki/\%E0\%A4\%AE\%E0\%A5\%8B\%E0\%A4\%A4\%E0\%A4\%BF\%E0\%A4\%B0\%E0\%A4\%BE\%E0\%A4\%AE_\%E0\%A4\%AD\%E0\%A4\%9F\%E0\%A5\%8D\%E0\%A4\%9F}{\texttt{Motiram Bhatta}} & People \\
ne.wikipedia.org&\href{http://ne.wikipedia.org/wiki/\%E0\%A4\%89\%E0\%A4\%AA\%E0\%A4\%A8\%E0\%A5\%8D\%E0\%A4\%AF\%E0\%A4\%BE\%E0\%A4\%B8_\%E0\%A4\%A4\%E0\%A4\%B0\%E0\%A4\%99\%E0\%A5\%8D\%E0\%A4\%97\%E0\%A4\%BF\%E0\%A4\%A3\%E0\%A4\%BF_\%28\%E0\%A4\%AA\%E0\%A4\%A4\%E0\%A5\%8D\%E0\%A4\%B0\%E0\%A4\%BF\%E0\%A4\%95\%E0\%A4\%BE\%29}{\texttt{Novel Magazine}} & Art/Culture \\
ne.wikipedia.org&\href{http://ne.wikipedia.org/wiki/\%E0\%A4\%AD\%E0\%A5\%8D\%E0\%A4\%B2\%E0\%A4\%BE\%E0\%A4\%A6\%E0\%A4\%BF\%E0\%A4\%AE\%E0\%A4\%BF\%E0\%A4\%B0_\%E0\%A4\%AA\%E0\%A5\%81\%E0\%A4\%9F\%E0\%A4\%BF\%E0\%A4\%A8}{\texttt{Vladimir Putin}} & Politics \\
ne.wikipedia.org&\href{http://ne.wikipedia.org/wiki/\%E0\%A4\%A8\%E0\%A5\%87\%E0\%A4\%AA\%E0\%A4\%BE\%E0\%A4\%B2\%E0\%A5\%80_\%E0\%A4\%B8\%E0\%A4\%BE\%E0\%A4\%B9\%E0\%A4\%BF\%E0\%A4\%A4\%E0\%A5\%8D\%E0\%A4\%AF\%E0\%A4\%95\%E0\%A5\%8B_\%E0\%A4\%87\%E0\%A4\%A4\%E0\%A4\%BF\%E0\%A4\%B9\%E0\%A4\%BE\%E0\%A4\%B8}{\texttt{History of Nelali Literature}} & History \\
ne.wikipedia.org&\href{http://ne.wikipedia.org/wiki/\%E0\%A4\%86\%E0\%A4\%AF\%E0\%A4\%95\%E0\%A4\%B0}{\texttt{Income tax}} & Law \\
ne.wikipedia.org&\href{http://ne.wikipedia.org/wiki/\%E0\%A4\%B0\%E0\%A4\%B5\%E0\%A4\%BF_\%E0\%A4\%AA\%E0\%A5\%8D\%E0\%A4\%B0\%E0\%A4\%BE\%E0\%A4\%9E\%E0\%A5\%8D\%E0\%A4\%9C\%E0\%A4\%B2}{\texttt{Ravi Prasjal}}$^+$ & People \\
ne.wikipedia.org&\href{http://ne.wikipedia.org/wiki/\%E0\%A4\%AF\%E0\%A5\%8B\%E0\%A4\%97\%E0\%A4\%9A\%E0\%A5\%82\%E0\%A4\%A1\%E0\%A4\%BE\%E0\%A4\%AE\%E0\%A4\%A3\%E0\%A4\%BF_\%E0\%A4\%89\%E0\%A4\%AA\%E0\%A4\%A8\%E0\%A4\%BF\%E0\%A4\%B7\%E0\%A4\%A6}{\texttt{Yogchudamani Upanishads}}$^+$ & Religion \\
ne.wikipedia.org&\href{http://ne.wikipedia.org/wiki/\%E0\%A4\%B8\%E0\%A5\%87\%E0\%A4\%A2\%E0\%A4\%BE\%E0\%A4\%88}{\texttt{Sedai}}$^+$ & Religion \\
\bottomrule

    \end{tabular}
    \caption{List of documents by Wikipedia domain,  their document name or English translation, and corresponding topics. The document name has an hyper-reference to the original document. $^{+}$ denotes a page that has been removed or no longer available at the time of this submission.}
    \label{tab:wiki_docs}
\end{table*}

\clearpage
\section{Examples from {\em devtest}}
\begin{table*}[!h]
    \centering
    \includegraphics{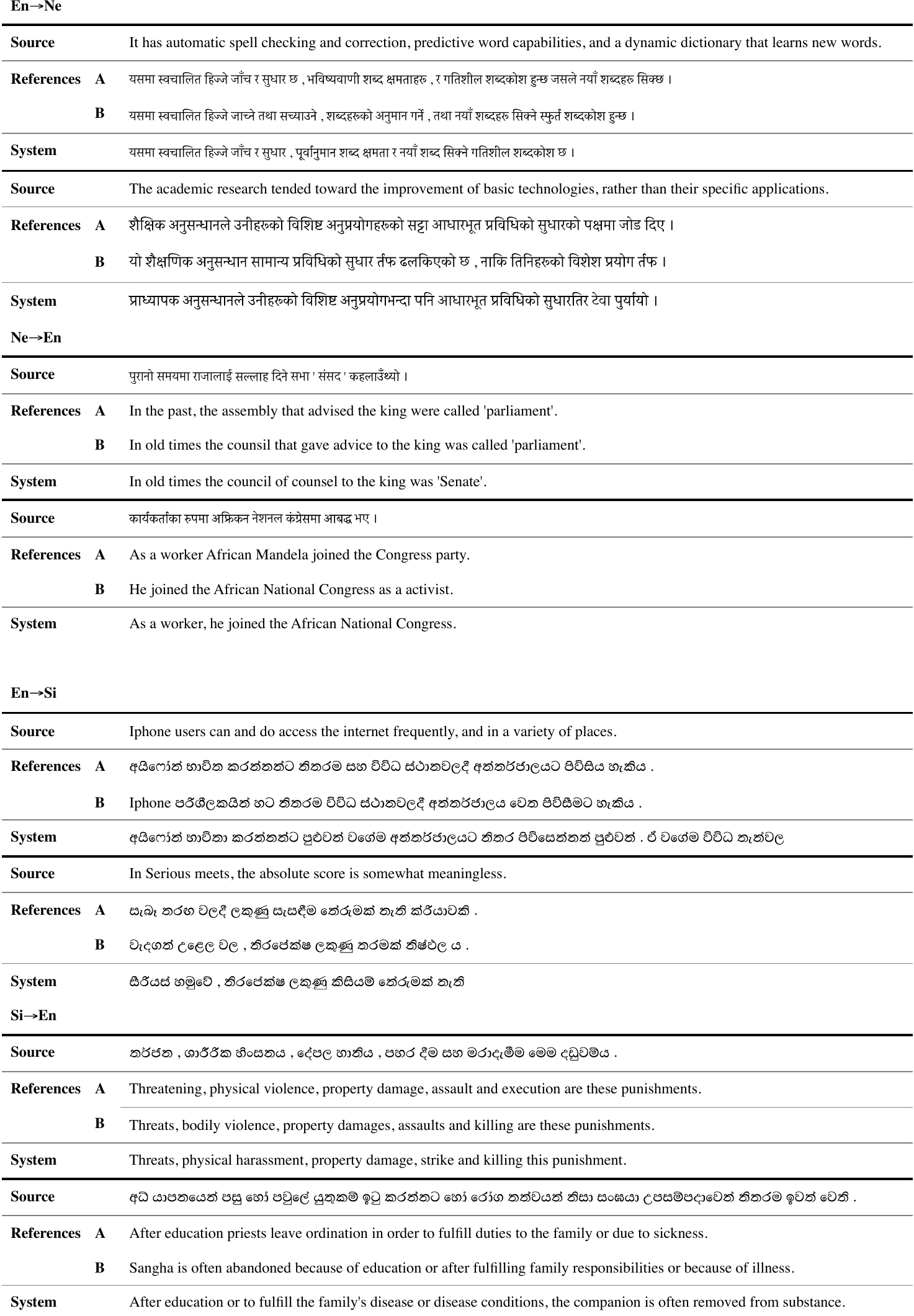}
    \caption{\small Examples of sentences from the En-Ne, Ne-En, En-Si and Si-En {\em devtest} set. System hypotheses (System) are generated using the semi-supervised model described in the main paper using beam search decoding.}
    \label{tab:examples}
\end{table*}

\end{document}